\documentclass[letterpaper, 10 pt, conference]{ieeeconf}  

\IEEEoverridecommandlockouts                              

\usepackage[dvips]{color}
\usepackage[noadjust]{cite}
\usepackage{epsfig}
\usepackage{amstext}
\usepackage{amsmath}
\usepackage{graphicx}
\usepackage{amsfonts}%
\usepackage{amssymb}
\usepackage{psfrag}
\usepackage{url}
\usepackage{xcolor}
\DeclareGraphicsExtensions{.pdf,.png,.jpg,.gif,.eps}
\graphicspath{{./figures/}}
\usepackage[official]{eurosym}
\usepackage{algorithm}
\usepackage{algpseudocode}
\usepackage{lipsum}
\usepackage{subfigure}

\usepackage{listings}
\usepackage{color}
\definecolor{codegreen}{rgb}{0,0.6,0}
\definecolor{codegray}{rgb}{0.5,0.5,0.5}
\definecolor{codepurple}{rgb}{0.58,0,0.82}
\definecolor{backcolour}{rgb}{1,1,1}
 
\usepackage{lipsum}
\usepackage{courier}

\lstdefinestyle{mystyle}{
    backgroundcolor=\color{backcolour},   
    commentstyle=\color{codegreen},
    keywordstyle=\color{magenta},
    numberstyle=\tiny\color{codegray},
    stringstyle=\color{codepurple},
    breakatwhitespace=false,                         
    captionpos=b,                    
    keepspaces=true,                 
    numbers=left,  
    xleftmargin= 15 pt,                 
    numbersep= 5 pt,                  
    showspaces=false,                
    showstringspaces=false,
    showtabs=false, 
    numberstyle=\fontsize{7}{9}\ttfamily,     
}

\newcommand{\figone}[4]{
\begin{figure}[tpb]
\begin{center}
\includegraphics[width=#4in]{#2}
\begin{sl}
\caption{\label{#1}#3}
\end{sl}
\end{center}
\end{figure}}

\IEEEoverridecommandlockouts
\begin{document}

\title{\LARGE \bf Node Primitives: \\ an Open End-user Programming Platform for Social Robots}
\author{
	Enrique Coronado, Fulvio Mastrogiovanni, Gentiane Venture%
	\thanks{
		Enrique Coronado and Gentiane Venture are with the Department of Mechanical Systems Engineering, Tokyo University of Agriculture and Technology, 2-21-16 Nakacho, Koganei, Tokyo, Japan.  Fulvio Mastrogiovanni is with the Department of Informatics, Bioengineering, Robotics and Systems Engineering, University of Genoa,
		Via Opera Pia 13, 16145 Genoa, Italy.
		{Corresponding author's email: enriquecoronadozu@gmail.com}.
	}%
}
 \IEEEoverridecommandlockouts

\maketitle

\begin{abstract} 
With the expected adoption of robots able to seamlessly and intuitively interact with people in real-world scenarios, the need arises to provide non technically-skilled users with easy-to-understand paradigms for customising robot behaviours. In this paper we present an interaction design robot programming platform for enabling multidisciplinary social robot research and applications. This platform is referred to  Node Primitives (NEP) and consists of two main parts. On the one hand, a ZeroMQ and Python-based distributed software framework has been developed to provide
inter-process communication and robot behaviour specification mechanisms. On the other hand, a web-based end-user programming (EUP) interface has been developed to allow for an easy and intuitive way of programming and executing robot behaviours. In order to evaluate NEP, we discuss the development of a human-robot interaction application using arm gestures to control robot behaviours. A usability test for the proposed EUP interface is also presented.


\end{abstract}
\section{Introduction}


Social robotics is a novel robot design paradigm with the potential of changing the way we interact with machines in our everyday-life. A number of research and technological challenges must be faced to introduce these robots in everyday environments \cite{venture2016personalizing}. One of the main obstacles to the adoption of social robots can be seen in the application point of view. This is due to the fact that many users of social robots are interested in designing new interactive applications involving both humans and robots. These users can be classified in \textit{robot software developers} and \textit{interaction designers} \cite{barakova2013end}. On the one hand, robot software developers are mainly interested in rapidly prototyping interactions and in integrating new algorithms for human behaviour understanding, dialog management, as well as emotion and body expression, among others. Users in this category typically include engineering students and researchers. On the other hand, interaction designers are mainly interested in the creation of new real-world applications or in the evaluation of theories about  human-robot interaction (HRI). Examples of this type of users include: educators, sellers, scholars and practitioners in social sciences and psychology, just to name a few. In most cases, interaction designers are the most qualified to design the associated robot behaviours and the interaction flows, given their expertise and domain-specific knowledge \cite{glas2012interaction}.  Generally, research and integration activities in HRI require interdisciplinary work and expertise to address all the challenges related to \textit{robot} and \textit{human factors}. The application of social robots in everyday scenarios cannot assume technical expertise for interaction designers.  Hence, interaction designers need to rely upon robot software developers and end-user programming (EUP) tools to generate new applications. However, available expert and user-friendly robot programming platforms present either usability or flexibility issues, such as, steep learning curve, selection of a wrong abstraction level for software development, cumbersome software installation, lack of code-reuse capabilities, difficulty in generating complex dynamic behaviours, among others \cite{diprose2016designing, barakova2013end, berenz2014targets}. Furthermore, many of these systems are not robot hardware or operating system neutral. These problems make the generation of new complex interactive applications with social robots a very tedious and time-consuming task. Therefore, the problem we address in this paper is how to design an easy-to-use and intuitive robot programming platform targeted at non-technically skilled users allowing for the creation of human-robot interaction scenarios inspired by social robotics considerations.  The hypothesis of our work is that, in order to increase the flexibility and usability of social robots it is important to create not only `easy-to-use", but also ``easy-to-develop"  and ``easy-to-customize" frameworks allowing users to modify built-in applications and to develop completely new ones.

 
In an effort towards the development of usable and flexible tools for social robots, we propose a human-centered robotic programming platform,  which we call Node Primitives (NEP). This platform is characterized by the design principles of usability, modularity and reusability of distributed robot software systems, and is inspired by a number of theoretical considerations about behaviour abstractions discussed in \cite{diprose2016designing}. 

The NEP platform is composed of a back-end and a front-end component. On the one hand, the back-end is intended to allow for an easy development of reusable software components: it provides inter-process communication mechanisms, supports the creation of reusable software modules and adopts a human-centered Application Programming Interface (API) for the specification of robot social behaviours. On the other hand, a front-end web-based interface adopting  Google Blockly technologies provides an end-user interface, , which can be connected to a variety of low-cost motion sensors to enable complex interactions with social robots. It also offers an easy and intuitive way to develop complex robot behaviours using a drag-and-drop approach where blocks (representing primitive robot behaviours) can be easily connected together. The front-end is intended to be the primary means for users to create and execute their own applications.  

In order to evaluate the main characteristics of NEP, we discuss how it has been used to develop gesture-based human-robot interaction applications using a NAO robot. A usability test has been performed in order to evaluate the proposed end-user interface.  

This paper is organized as follows. In Section II, we discuss the related work. Section III describes the NEP robotic programming platform. Section IV presents the an experimental assessment. Conclusions follow.

\section{Related Work}

\subsection{Social Robot Programming Frameworks}
An analysis of available literature shows that two options are available to develop interactive applications for social robot: the use of expert-oriented programming frameworks and user-friendly  programming interfaces. In most of the available expert-oriented programming frameworks, the main concern is performance rather that usability. Therefore, they are generally difficult to install, use and learn. Moreover, they are based on GNU/Linux systems, which is not typically  used by  novice users \cite{barakova2013end}. 
These issues make the cooperation between robot software developers and interaction designers difficult. In fact, most of the frameworks available in the community do not provide interfaces targeted at interaction designers, nor can they be used without expert knowledge. Examples of such frameworks are BONSAI \cite{siepmann2011modeling} and the Robot Behavior Toolkit \cite{huang2012robot}. A number of user-friendly programming interfaces have also been proposed recently. Examples are Interaction Composer \cite{glas2012interaction}, Choregraph  \cite{pot2009choregraphe}, TiViPE \cite{lourens2011user} and Interaction Blocks \cite{sauppe2014design}. However, the majority of these solutions are neither \textit{robot neutral} nor oriented towards code reuse and modularity. Furthermore, from a practical perspective, some of them cannot be accessed or obtained easily, need computationally demanding third-party software or lack documentation resources for developers. 

\subsection{Correct Abstraction Level for Social Robot Programming}
\label{se:correct}
According to \cite{diprose2016designing}, \cite{green1996usability}, \cite{kolling2016heuristic}, it is important to choose a correct programming abstraction level when developing programming tools for novice users.  An analysis of the most appropriate programming abstraction level for social robotics APIs have been performed by Diprose and colleagues in \cite{diprose2016designing}. They propose five abstraction levels composed of heterogeneous primitives, which are described below:


\begin{itemize}
\item \textit{hardware primitives} are the low programming abstraction level, which is used to control robot actuators or acquire data from sensory devices;
\item \textit{algorithmic primitives} provide the robot with the abilities to \textit{make sense} and interact with their environment; these primitives are related to the typical robot perception tasks including (as far as social robots are concerned), speech recognition, gesture recognition and face tracking; 
\item \textit{social primitives} corresponds to a domain-specific robot behaviour generally implemented with algorithmic and hardware primitives; a possible example is a gaze control where the robot needs to combine face tracking and head movement control; 

\item \textit{emergent primitives} are high-level abstractions, which are implemented  combining a number of social primitives; a possible example is a human-robot dialogue where the robot needs to combine social primitives such as gaze control, speech-based utterances and body gestures; 
\item \textit{control primitives}  are used to orchestrate (i.e., synchronise or sequence) the other primitives and to generate actual robot behaviours; examples of control primitives are rule-based systems, finite-state machines or action planning frameworks. 
\end{itemize}

The authors of \cite{diprose2016designing} argue that most of the expert-oriented and novice robot programming tools present usability or flexibility issues due to the mixing of different types of primitives or the use of only high-level emergent primitives in the development of the related APIs. They also argue that high quality, flexible and usable social robot APIs must be composed as much as possible of domain-specific social primitives for behaviour specification. The NEP programming platform has been designed to consider the insights put forth by Diprose and colleagues as a functional requirements.

\section{Node Primitives as a Robot Software Development Platform}

\subsection{Design Principles}

The design principles which have inspired the NEP system?s architecture  are related to its \textit{usability}, \textit{flexibility}, being \textit{multi-paradigm} and \textit{open source}.
We discuss each principle and its contributions to NEP in the following paragraphs.


\textit{Usability}.
In NEP, usability issues are considered at different levels, namely software development, the definition of an API strongly based on the concept of social primitives, and the creation of a graphical user interface to hide technical aspects in robot behaviour development. Despite recent efforts to uniform robot software development according to a number of standards \cite{bischoff2010brics}, the adoption of freely available software components and frameworks is still difficult as far as installation, configuration and required expertise is concerned.  These tasks, when carried out by novice users (and in general by non technically skilled people) constitute a severe entry-level adoption barrier. In the current robot software development \textit{panorama}, usability means \textit{cross-platform}, \textit{easy to install}, \textit{learn}, \textit{develop} and \textit{distribute}.  In NEP, we decided to adopt Python as the reference programming language. Python is available for a wide selection of operating systems and can be also used for embedded computing.  Python focuses on writing readable, simple, explicit and clean code and is widely acclaimed by it short learning curve. It is also provided with a number of excellent and easy-to-use package installation and management tools such as \textit{Pip Installs Packages} (PIP) and \textit{setuptools}. 

\textit{Flexibility}. In NEP, we support the idea of distributing the basic skills (primitives) of the robot software into independent and interchangeable software components rather structuring the robot applications as a single component.  Inspired by the information processing cognitive model described in \cite{wickens2015engineering}, 
we classify software components in NEP as belonging to four different \textit{node} types:

\begin{itemize}
\item a \textit{sensory node} performs data acquisition, and collects raw information from different sensors to publish relevant data using a well-defined set of data types; this kind of node implements the Sensor/Device design pattern \cite{kahn2008design};

\item a \textit{perceptual node}  processes the data published by a sensory node to obtain numerical, sub-symbolic or symbolic representation structures; algorithmic  primitives are defined in perceptual nodes; several perceptual  nodes can be concatenated, therefore implementing the so-called Computational design pattern \cite{kahn2008design}; however, the output of the last node in the chain node must clearly define a representation of the robot's, human's or environment's state;
\item a \textit{cognitive node} orchestrates the overall behaviour of a set of nodes;  the development of cognitive nodes is where interaction design occurs; in particular, cognitive nodes define the human-robot interaction model or \textit{flow} \cite{kahn2008design};
\item an \textit{action node} represents robot behaviours and, on the basis of the interaction flow defined in relevant cognitive nodes, selects how to syncronise or sequence them to generate actual robot actions. 
\end{itemize}

\textit{Multi-paradigm network architecture}.
Distributed systems are characterised by a number of trade-offs as far as efficiency, determinism, scalability, security and robustness are concerned \cite{zero}. In NEP, we support multiple network architectures, each one with specific features. In the current version of our framework, fixed endpoint descriptions, central services and a Peer to Peer (P2P) server are supported. Figure \ref{fig:master} shows the adopted service discovery approach. This approach is a ROS-like P2P name server denoted as NEP Master.  Publisher and Subscriber instances implicitly connect to the NEP Master node using a  Client-Server protocol to perform registration requests for message topics. When the NEP Master detects a topic registration request, assigns a specified topic with an endpoint direction (i.e., an IP address and a port). This endpoint information is saved and sent to the Publisher or Subscriber instances which have performed the registration request of that topic. Communication starts when at least one Publisher and one Subscriber have requested the registration of the same topic. 

\figone{fig:master}{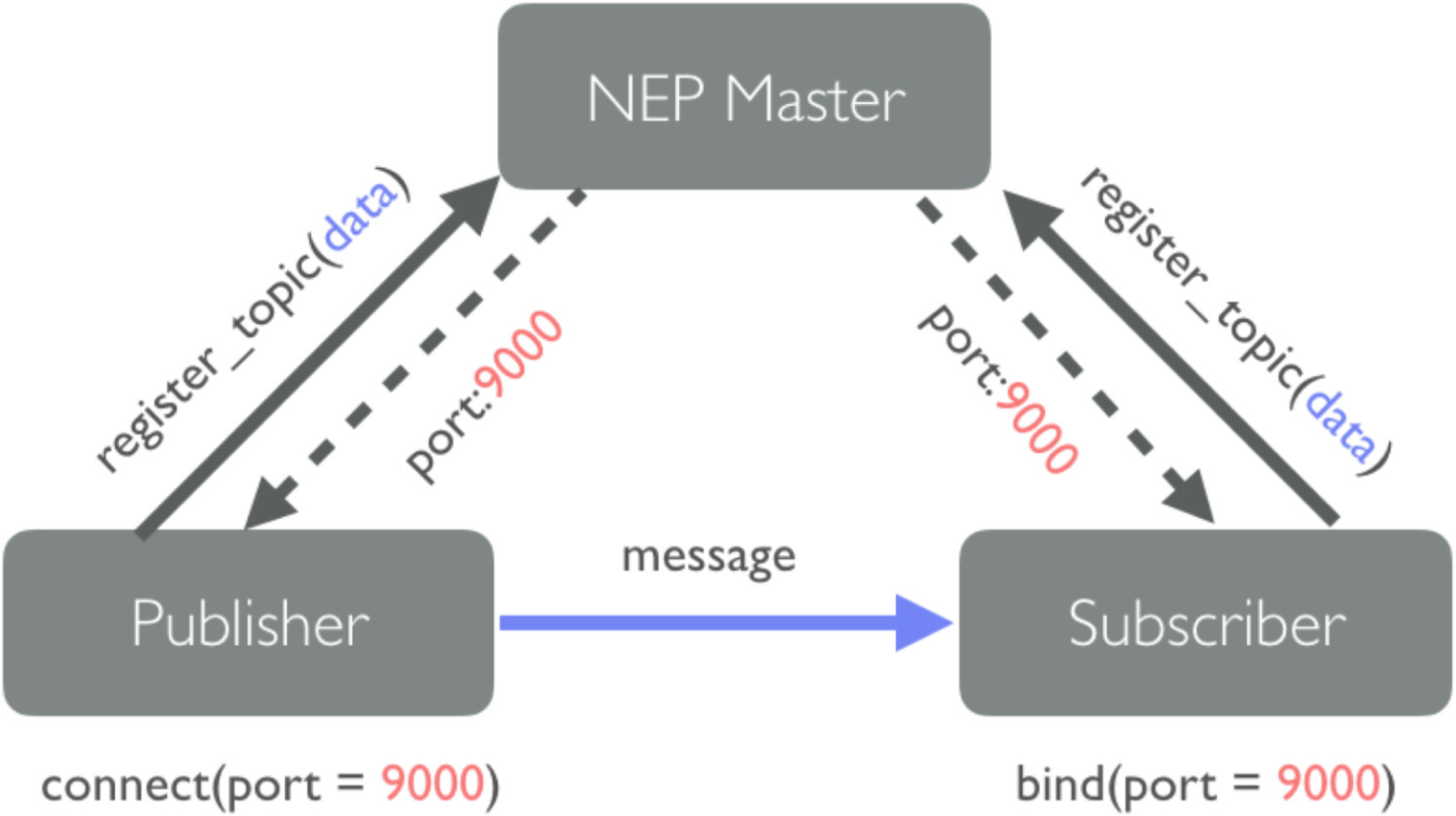}{A representation of service discovery and Publish-Subscribe communication.}{3.1}

\textit{Open source}. As done in other projects in robot software development, we support the idea of free and open source software. NEP is released under the MIT license \cite{open2006license}. 



\subsection{The NEP API}
\label{se:api}

In order to support  common robotic programming tasks such as inter-process communication, robot behaviour definitions and the node launching process, the NEP API provides a number of supporting nodes and programming interfaces (in terms of classes and data structures). 

\textit{The communication module}. This module provides all the interfaces and mechanisms described above to make all nodes communicate with each other. It uses the ZeroMQ library \cite{zero} to allow for a Publish-Subscribe inter-process communication between nodes. The current version of NEP supports communication between nodes developed in Python. However, nodes can be developed in other programming languages supporting the ZeroMQ library, such as C++, C\#, Javascript, among others. In order to make data formats associated with different languages compatible,  messages can be encoded and decoded using JSON \cite{crockford2006application} or a simple string format. The communication module is composed of two main classes: the $\mathsf{publisher}$ class and the $\mathsf{subscriber}$ class. These classes are aimed at hiding the complexity associated with  low-level ZeroMQ details. Let us consider an example where a perceptual node detects simple human emotions using visual information and provides a cognitive node with labels about perceived emotions. This can be implemented using a Publish-Subscribe mechanism where the perceptual node embeds a Publisher and the cognitive node employs a Subscriber. The two nodes communicate over a topic called $\mathsf{human\_behaviour}$, as follows:

\begin{lstlisting}[language=Python]
# Define new publisher
pub = publisher("/human_behavior")
# Define the message to send
msg = {'human_state': "happy"}
# Send the message
pub.send_info(msg)
\end{lstlisting}

In this example,  the \textit{topic} denoted as $\mathsf{human\_behavior}$ is defined in the new Publisher instance $\mathsf{pub}$ to enable topic filtering. In NEP, messages must be defined as strings or Python dictionaries. A Python dictionary is an unordered set of pairs $key:value$, separated by commas.  A key can be a string or a number and it is used as an index to save or extract values of any Python supported type. In the example above the dictionary denoted as $\mathsf{msg}$ is composed of a pair $\mathsf{'human\_state':``happy"}$.  Finally, the $\mathsf{send\_info()}$ method sends $\mathsf{msg}$ using JSON serialization. An example of a code snippet creating a $\mathsf{subscriber}$ instance and reading a message is: 

\begin{lstlisting}[language=Python]
# Define new subscriber
sub = subscriber("/human_behavior")
# Read a message in blocking mode
success,msg = sub.listen_info(block=True)
human_state = msg["human_state"]
\end{lstlisting}

In this example a cognitive node listens the messages published in the $\mathsf{human\_behaviour}$ topic. The read action can be performed in blocking or non blocking mode. The selection of this mode is defined by the $\mathsf{block}$ argument in the $\mathsf{listen\_info()}$ function. The default value of this parameter is blocking mode. In the non-blocking mode the $\mathsf{success}$ return parameter  indicates whether data have been received under a timeout. The $\mathsf{msg}$ return parameter contains the message  deserialized as a Python dictionary.  In the example above the value of the message is extracted from the Python dictionary using the corresponding key index denoted as $\mathsf{'human\_state'}$.

\textit{The launcher class}. Using this class it is possible to launch nodes written in Python. In order to launch a node, it is necessary to specify the node primitive type (i.e., sensing, perception, cognitive or action) and the node's name. Other parameters such as the IP  address associated with the robot in the network can be provided as well. For example, launching a perceptual node called $\mathsf{gesture\_recognition}$ is achieved as:

\begin{lstlisting}[language=Python]
# Basic defintion of a node
type_node = "perceptual"
node_name = "gesture_recognition"
# A new launcher instance
lan = launcher()
# Launch the node
lan.nep_launch(type_node, node_name)
\end{lstlisting}

\textit{The robot class}. This class defines a set of social robotics oriented behaviours, typically used in cognitive nodes. All implemented methods are conceptually based on the abstraction level suggested in \cite{diprose2016designing}. As an example, in order to create a $\mathsf{robot}$ instance $\mathsf{r}$, link it to a real-world NAO robot, and command NAO to perform a sequence of basic behaviours (namely, say something without any associated gesture, change posture and imitate a cat), it suffices to write the following code:

\begin{lstlisting}[language=Python]
# Create a new robot instance
r = robot()
# Select the robot to use
r.setRobots(["NAO"])

# Text to speech request
r.say("Hello", gesture = "no_gesture")
# Robot posture request
r.do("posture_stand")
# Robot imitation request
r.imitate("cat")
\end{lstlisting}

In order to execute all robot behaviours, the $\mathsf{robot}$ class creates a $\mathsf{publisher}$  instance, which sends the specification of the actions that the robot must execute to the relevant action node. Each action nodes receives and executes the behaviour specifications for the corresponding robot.  Multiple robots can perform the task in parallel. For this, a list of the relevant action nodes must be defined in  the $\mathsf{setRobots()}$ function. Each action node is differentiated by a robot name. 

\textit{The robot behaviour class}. This class is aimed at defining \textit{complex} (i.e., made up or simpler behaviours sequenced or executed in parallel) and \textit{dynamic} (i.e., reactive and adaptive) robot behaviours. In the current version of NEP, only reactive behaviours are supported, which are specified by simple \textit{if-then} rules. Rules specifying reactive robot behaviours are defined in a $\mathsf{reactive\_function()}$, which is associated with a given robot $\mathsf{behaviour}$, as follows: 
 
\begin{lstlisting}[language=Python]
# Creation of a reactive behavior
behavior = robot_behaviors("reactive") 
# Add a function to the behavior class
setattr(behavior, "reactive_function"
        ,reactive_function)
# Start robot behavior
behavior.run()

\end{lstlisting}

In this example the reactive function is added to the behaviour class using the Python built-in function $\mathsf{setattr()}$. Then, the behaviour is executed using the method run(). An example of a reactive function where the say ``Impressive!" each time a gesture denoted as $\mathsf{karate}$ is detected is shown bellow:

\begin{lstlisting}[language=Python]
# Example of a simple reactive function
def reactive_function(self,human_state):
    if human_state == "gesture_karate":
      r.say("Impressive!")
\end{lstlisting}

\subsection{The NEP End-user Programming Interface}
\label{se:in}

\figone{fig:server}{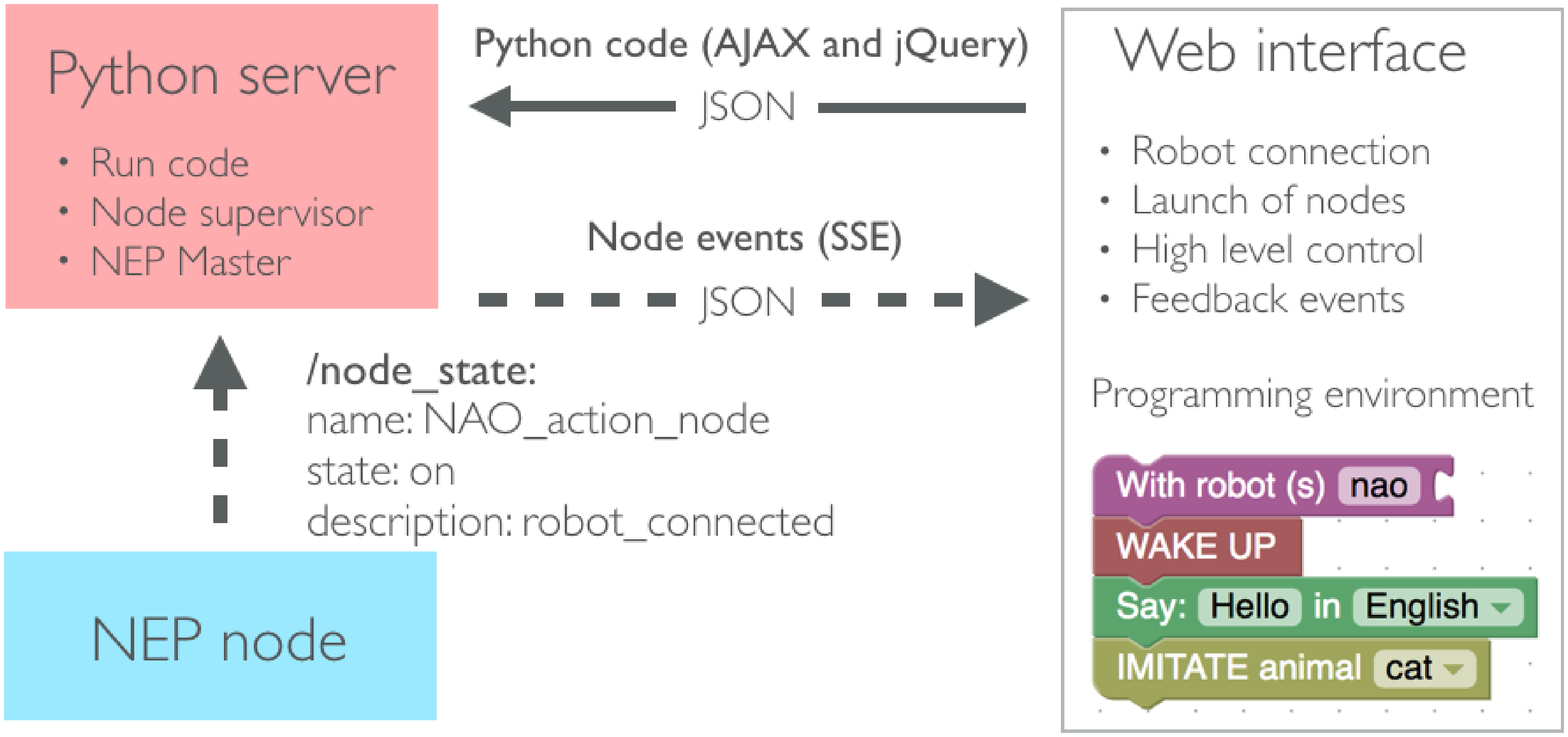}{Integration of NEP framework and the web-based interface.}{3}

The proposed high-level programming interface is based on a series of web-based, graphical tools designed to maximise usability as well as aesthetics. We select a block-based interface instead of a dataflow mechanism for a number or reasons, namely:

\begin{itemize}
\item dataflow approaches can lead to a closeness of mapping problem (i.e., elements in the notation do not clearly describe the corresponding objects in the formalism), which affects the understandability and maintainability of programs when reactive and dynamic behaviours are defined \cite{berenz2014targets}, \cite{green1996usability};
\item when the number of wire connections in a dataflow diagram increases, it becomes likewise more complex to understand the actual information flow \cite{berenz2014targets};
\item block-based approaches are becoming common tools used by educators and non specialists to get started with software development; furthermore, they are considered more engaging for novice users \cite{weintrop2015block}.
\end{itemize}

This high-level programming interface is designed as a Client-Server architecture (Figure \ref{fig:server}). The Client module provides the user interface, which is shown in Figure \ref{fig:inter}. This interface is composed of three block environments which enable end-users to define and sequence robot behaviours, connect robots in the network and launch the different nodes as part of the application, respectively. The code needed to perform these actions is embedded in each behavioral block.  
The code generated by the blocks can be downloaded or sent to the Server module using AJAX and jQuery asynchronous methods and JSON serialization for its execution.

A Python-based server allows for the access to the Client interface, executes and saves the received code, and provides feedback about the state of the various nodes. To this aim, each NEP node publishes messages on the topic  $\mathsf{node\_state}$,
which specifies the node name, the node type (i.e., sensory, perception, cognitive and action) and a description of a state event. The Server subscribes to this topic and uses Server-Sent Events (SSEs) to launch asynchronous events in the user interface. Examples of currently supported node events are: successful node start,  code in execution, correct or wrong connection with a robot, unexpected or manual node shutdown.  

\figone{fig:inter}{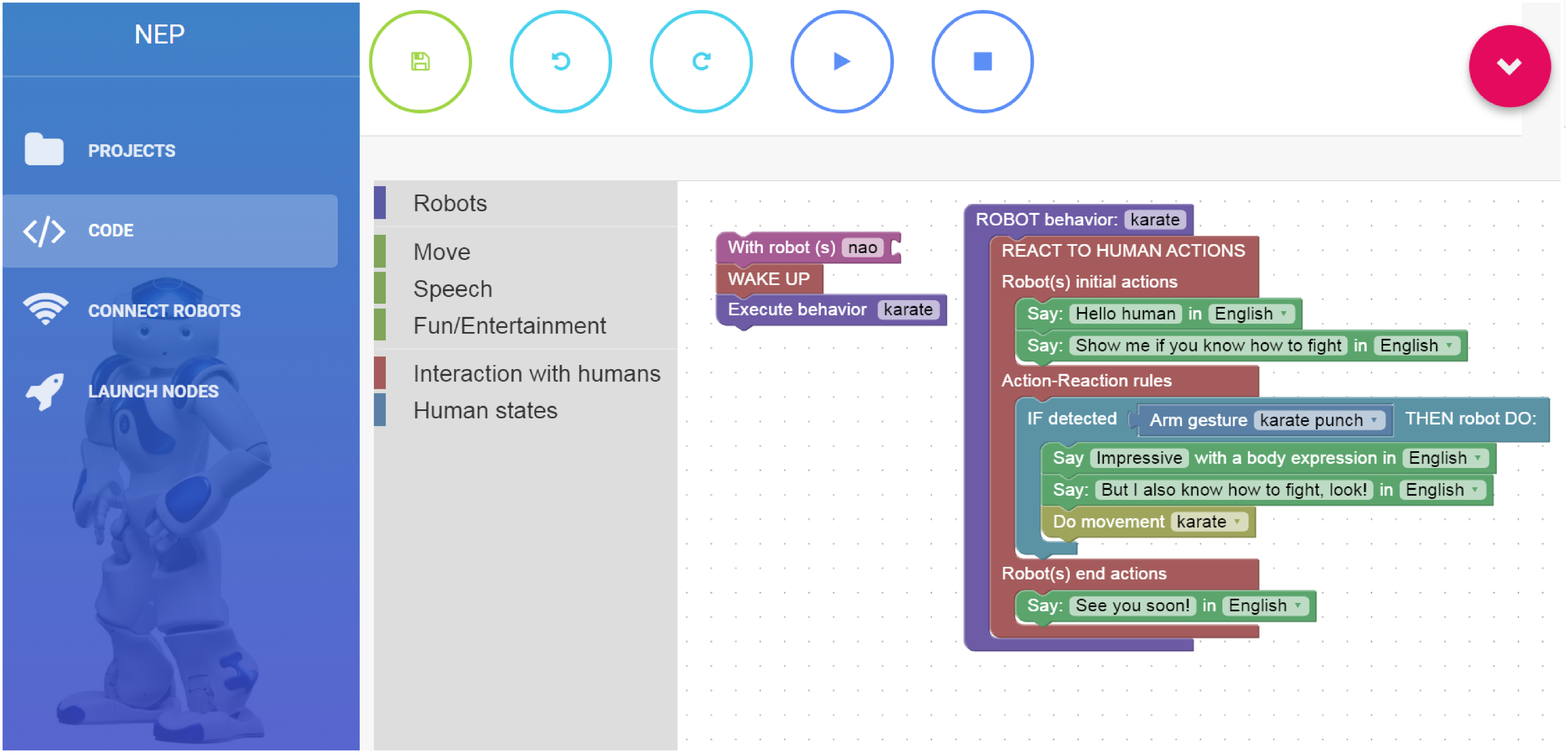}{End-user programming, node launching and behaviour execution interface developed}{3.3}
 
\section{Evaluation with NEP Early Adopters}

%
%
%
%
%
%
%
%
%
%

\subsection{Gesture-based Interactive Interaction with Social Robots}
\label{se:ge}
In order to test the operational capabilities of the NEP platform, we have developed an application allowing users to interact with NAO robots by means of gestures. Figure \ref{fig:gesture_ar} shows the application?s architecture. Sensory information is obtained by a smartwatch worn by the human interacting with the robot and sent via Wi-Fi. Gesture information is acquired by a sensory node and published 
the NEP communication module. A number of perceptual nodes encapsulating the approach proposed in \cite{coronado2017gesture}
can be used to detect and classify a set of predefined gestures. Figure
\ref{fig:all} shows snapshots of the gestures used in  this test. They are denoted as \textit{katana}, \textit{batting}, \textit{hand up}, \textit{karate}, \textit{strech up}. The system proposed in \cite{coronado2017gesture} assumes each gesture of interest be learned via Gaussian Mixture Modeling and Regression, and then classified online using statistical classification techniques \cite{bruno2013analysis}. For our tests, a set of $10$ recordings from $5$ volunteers have been collected to account for the variability in motion modeling. An \textit{ad hoc} benchmarking procedure to assess the recognition and classification performance of the system has been carried out, which is presented in Figure \ref{fig:kunfu}.
It can be observed that \textit{hand up} and \textit{karate} are in general always recognised, and good performance also characterises \textit{stretch up} and \textit{katana}. Performance worsens in case of \textit{batting}, which is occasionally mistaken for \textit{katana} and often not even recognised. A detailed analysis of this result is out of the scope of this paper. The interested reader is referred to \cite{bruno2013analysis}.

\figone{fig:all}{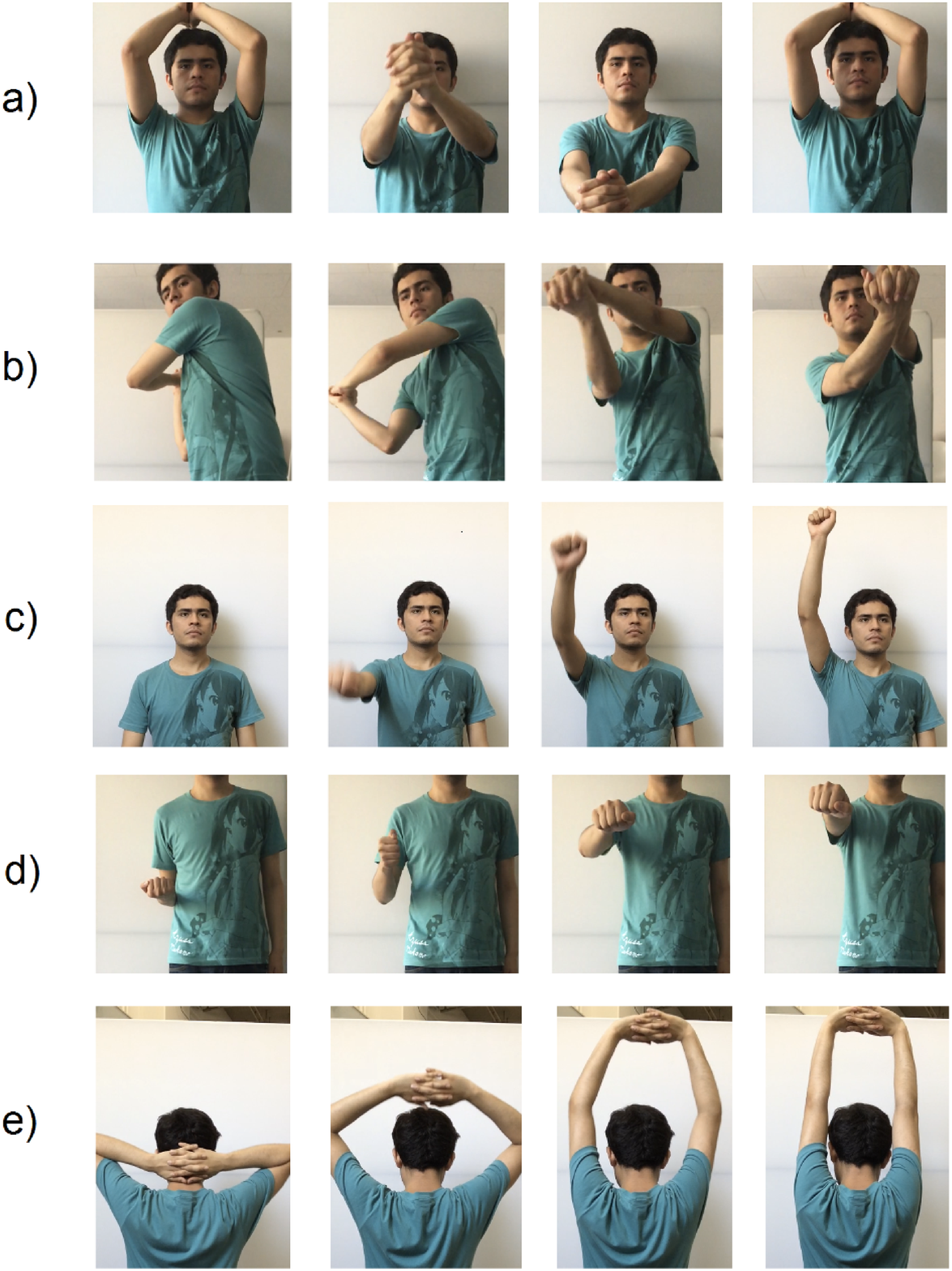}{Gestures recognized by our test application: a) \textit{katana} b ) \textit{batting} c) \textit{hand\_up} d) \textit{karate} e) \textit{strech\_up} }{2.7}

\figone{fig:kunfu}{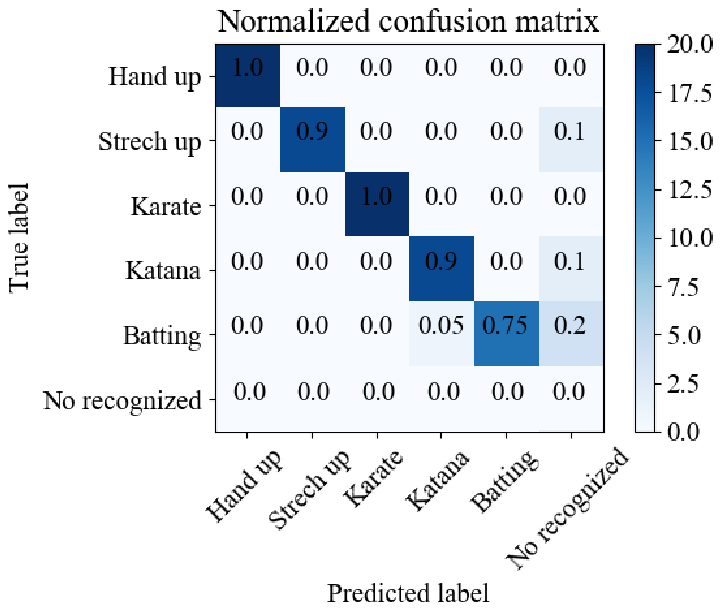}{Classification results for the five gestures of interest.}{2.9}

Each perceptual node is in charge of processing sensory data collected by the smartwatch and to look for a specific gestures. Therefore, five perceptual nodes are executed in parallel in our application. When a perceptual node detects a gesture, a trigger is raised (i.e., a JSON message using a NEP Publisher).  From these triggers, a cognitive node can infer and control the execution of the robot behaviours as defined by the user.
Behaviour specifications are published in order to be read and executed by an action node for a specific robot. In our case, a NAO specific action node has been developed and used. This node exploits the NAO SDK to connect via Wi-Fi to a NAO robot and executes the defined behaviours. Once a behaviour is executed, the action node publishes the result of the execution, which is read by the cognitive node to continue with the flow of the interaction. The execution, launching and supervision of all the aforementioned nodes can be performed by the Python-based Server and the web-based interface described in Section \ref{se:in}.

\figone{fig:gesture_ar}{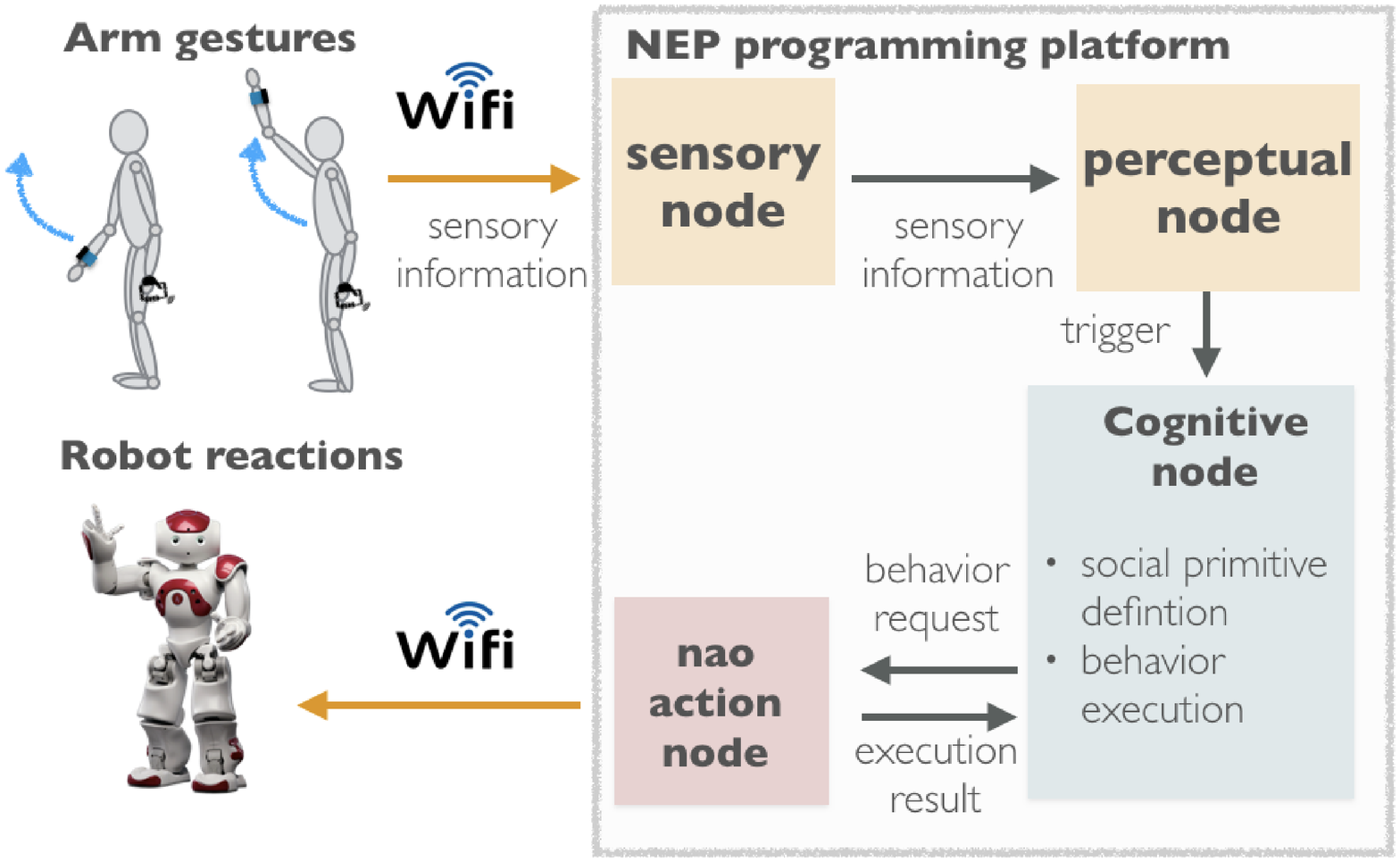}{The architecture of the gesture-based human-robot interaction test application. }{3}

As an example, Figure  \ref{fig:test0} shows a sequence of snapshots of a gesture-based application programmed in NEP, which follows the workflow described in Figure \ref{fig:test1}. The human first executes a \textit{karate} gesture. When it is detected, NAO reacts as specified by the program: it first says ``Impressive!" and then executes a scripted sequence of motions and dialogues.

\figone{fig:test0}{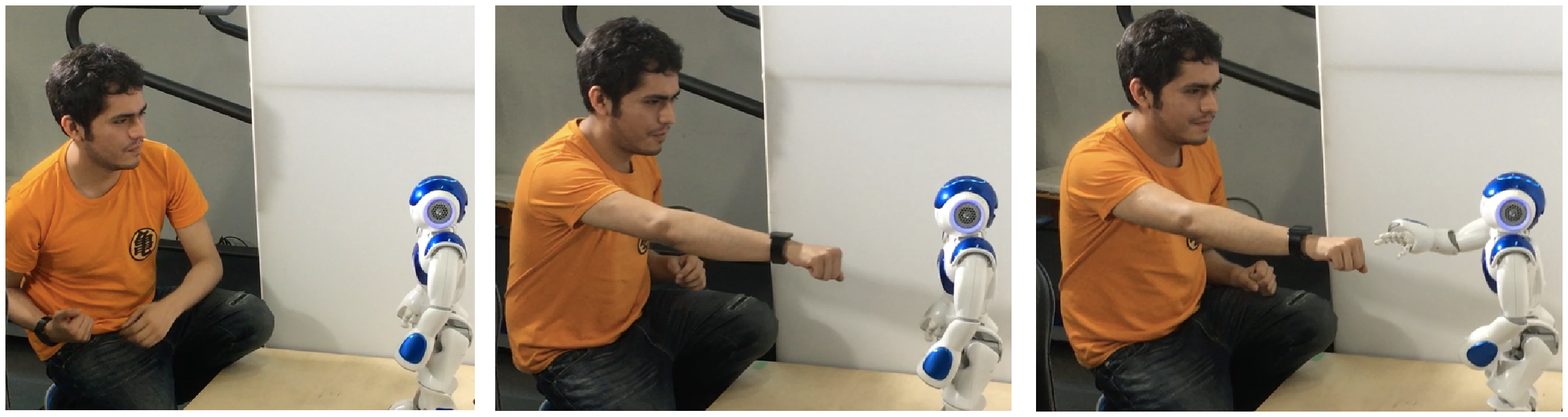}{ A NAO robot reacting to a \textit{karate} gesture.}{3.2}


%
%
%

\subsection{Evaluation of End-user Programming Interface} 
\label{se:usab}
In order to validate and improve the quality of the proposed end-user programming interface, an initial usability test with early adopters has been performed. Five participants have been involved. As suggested by the Nielsen Norman Group, this number is enough to discover an $80\%$ of the usability issues and the use of more participants can be a waste of resources \cite{nnwhy2017}. The subjects selected for these tests have different nationalities and, their programming skills range from none to expert-level. The tests started with a brief explanation of the objectives and activities to perform using a set of slides that also discuss the use of the main features of the interface. These slides also specify  the task to perform, i.e., creating an interactive application using the NAO robot and the gesture-based application described in the previous Section. After the programming task, the System Usability Scale (SUS) questionnaire was administered. The used SUS version has seven options (from strong disagree to strong agree) for each item. These items are alternated between positive and negative connotation. The scoring system of this test is somewhat complex. A detailed analysis of the scoring system is described in \cite{brooke1996sus}. Due to the fact that the participants were non English native speakers, a number of modifications to the SUS questionnaire have been done as suggested in \cite{finstad2006system}. The obtained SUS results (table \ref{tb:sus}) show that participants mostly have positive feelings about the interface. According to the acceptability span proposed in \cite{bangor2009determining}, the usability score obtained for the proposed interface is in the range of acceptability.

\figone{fig:test1}{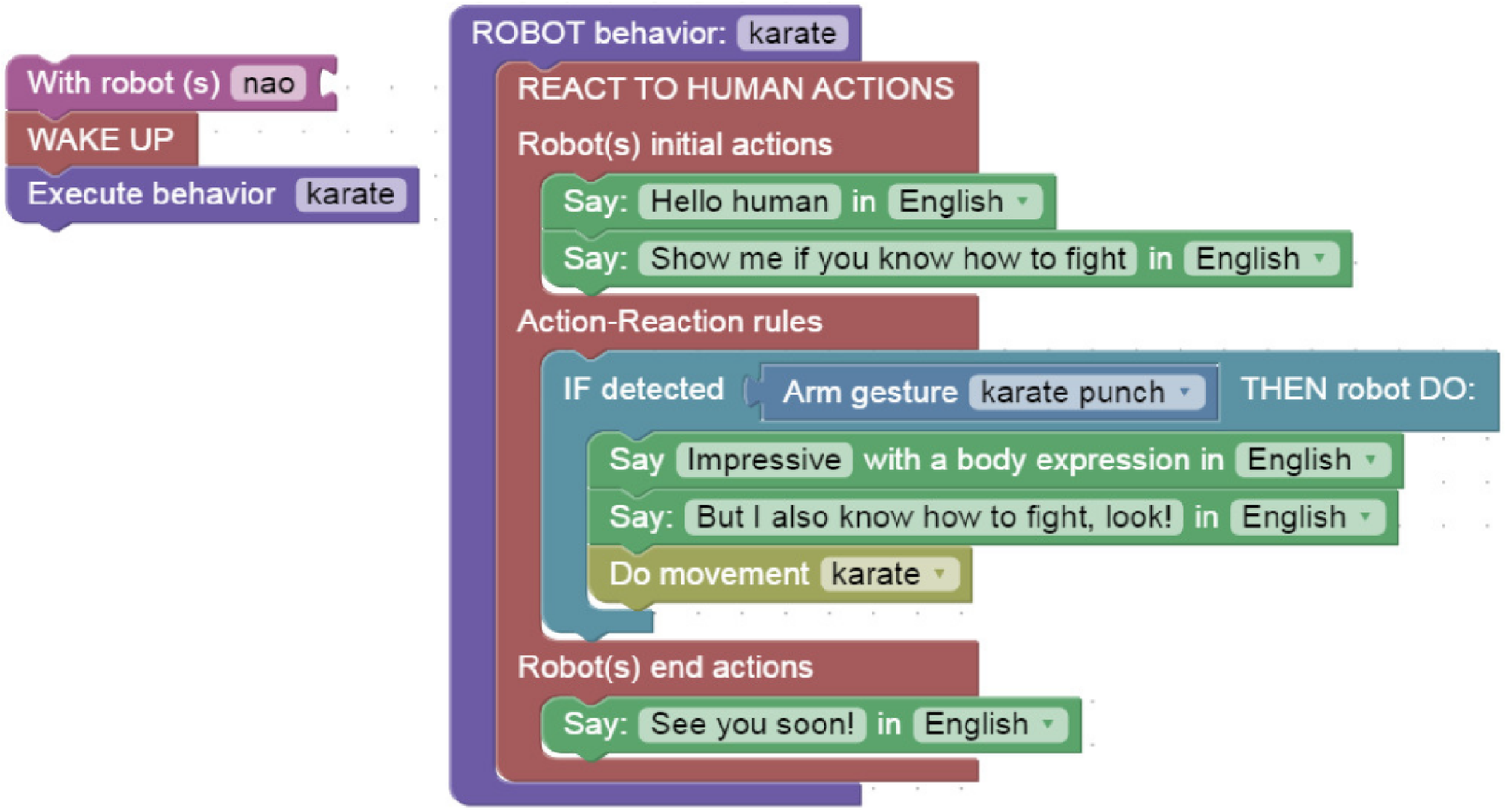}{A program developed to perform a simple human-robot interaction based on the five defined gestures.}{3.3}

\begin{table}[]
\centering
\caption{ Subjective Satisfaction Scores for the proposed interface}
\label{tb:sus}
\begin{tabular}{|l|l|l|p{1cm}}
\hline
\textbf{Question}                                                                            & $\mu$ & $\sigma$ \\\hline
I think that I would like to use this interface frequently                                   & 5.6     & 0.47   \\ \hline
I found the interface unnecessarily complex                                                  & 2.1     & 0.89   \\ \hline
I thought the interface was easy-to-use                                                      & 6.5     & 0.5    \\ \hline
I needed technical support to use this interface & 4       & 1.52                                                                                                                                               \\ \hline
I found the functions in this interface well integrated                         & 4.8     & 1.06   \\ \hline
There was too much inconsistency in this interface & 2.1     & 1.06   \\ \hline
Most people can learn to use this interface very quickly                        & 6.3     & 0.74   \\ \hline
The interface is very difficult/awkward to use                                       & 2       & 0.81   \\ \hline
I felt very confident using the interface                                                    & 5.5     & 0.76   \\ \hline
I learned a lot of things before I could use the interface                      & 1.5     & 0.5    \\ \hline
\multicolumn{3}{|c|}{\textbf{Overall SUS score (out of 100): 78.6}}                                              \\ \hline
\end{tabular}
\end{table}

Further recommendations to improve the usability of the interface, which emerged during these first trials, are:

\begin{itemize}
\item creating a \textit{reference section} in the interface to allow the user to look for documentation about block functionalities;
\item showing \textit{examples} of block applications;
\item showing \textit{help messages} related to how running a program;
\item add \textit{checklists} to error messages.
\end{itemize}

In general these recommendations are related to the documentation and help resources. All of these recommendations will be considered for the next version of NEP.


\section{Conclusions and Future Work}

In this paper, we present and discuss Node Primitives, a programming platform specifically targeted towards the development of software applications for social robots. 

Node Primitives has been designed according to a number of functional requirements and best practices, which have been discussed in the literature in the past few years. In particular, Node Primitives is divided in two main parts: while a back-end (oriented to robot software developers) organises a robot architecture as a set of nodes, each one characterised by a specific functionality, a graphical front-end (intended for interaction designers) allows for organising the human-robot interaction \textit{experience} using an intuitive block-based approach.

Node Primitives has been used already by non specialists in two scenarios: an interactive game for children in a museum (Figure 9) and to sequence robot actions in an experimental theatrical piece (Figure 10). The current version of Node Primitives is available open source\footnote{Web: \url{https://github.com/enriquecoronadozu/NEP}.}. 

Future work will be focused in:
\begin{itemize}
\item Supporting of deliberative and hybrid robot control behaviours and cognitive architectures
\item Development of a web-based simulator
\item Supporting of more social and service robots and low-cost sensory devices
\item Supporting of learning from demonstration and teleoperation tools 
\item Supporting of multi-robot architectures and enable code reuse with other robotic frameworks
\end{itemize}


\figone{fig:real}{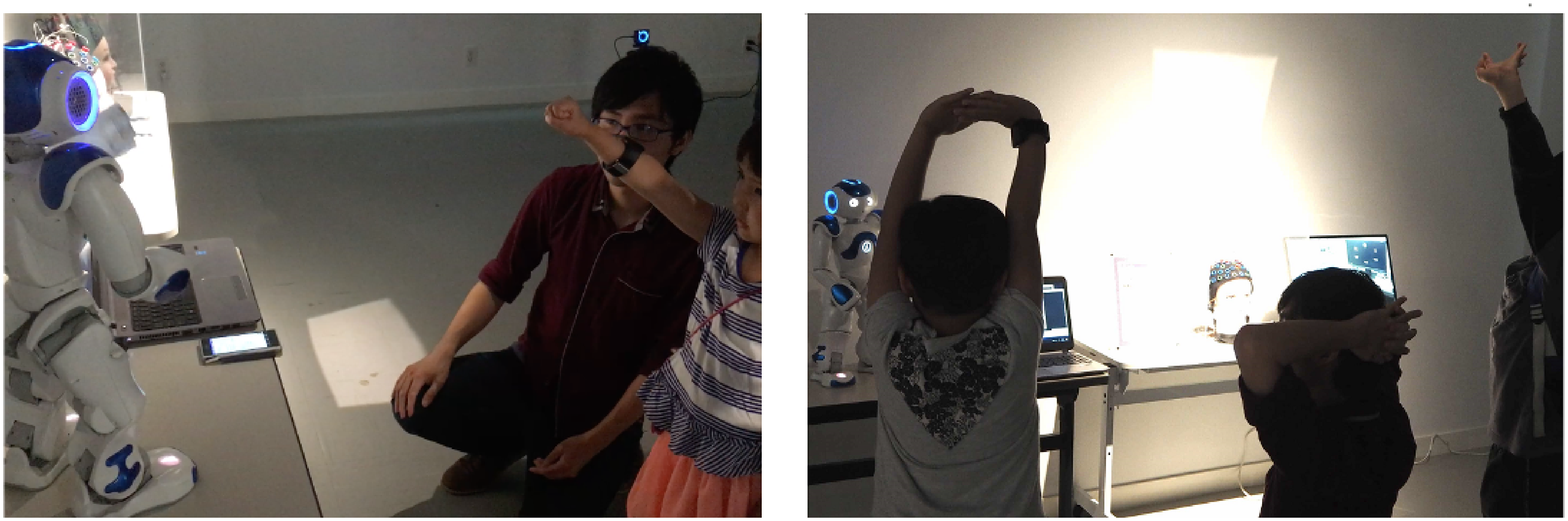}{Robot coach interactive application in museum}{3}
\figone{fig:co}{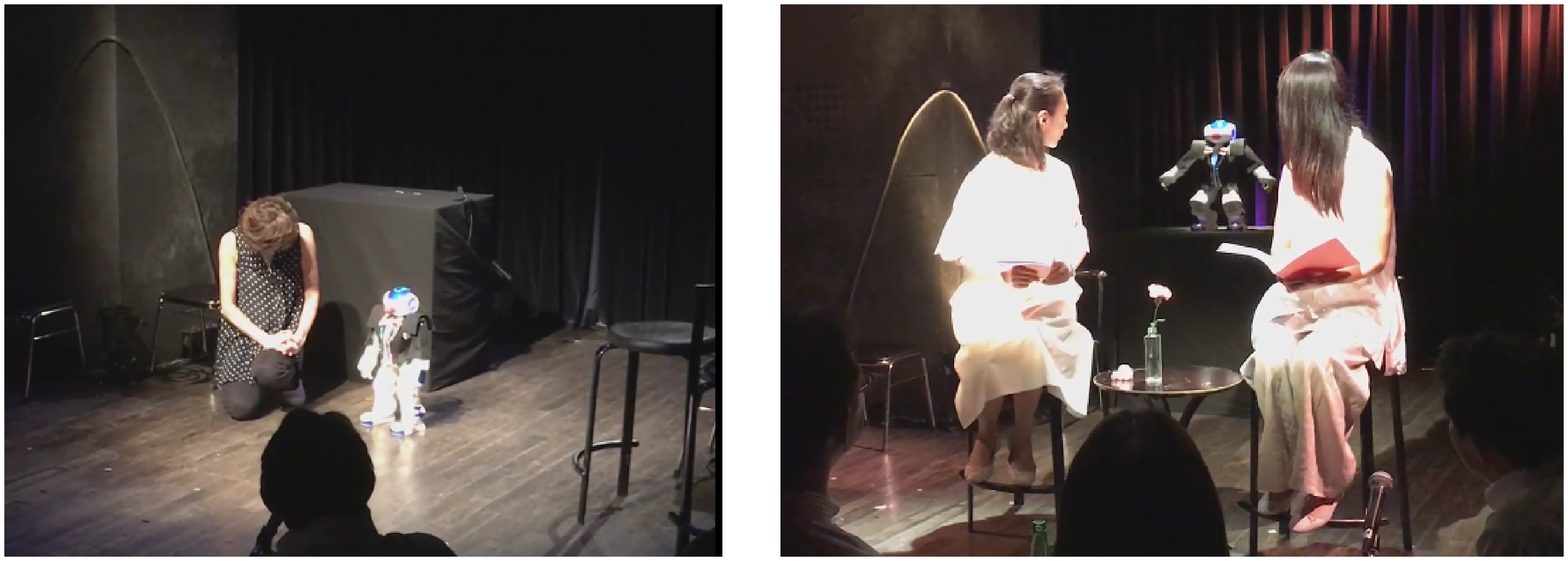}{NAO robot on stage in a theater presentation}{2.9}

\bibliographystyle{IEEEtran}
\bibliography{./social}

\end{document}